\documentclass[sigconf,acmtog,natbib]{acmart}

\AtBeginDocument{%
  }
\copyrightyear{2025}
\acmYear{2025}
\renewcommand\footnotetextcopyrightpermission[1]{} 

\usepackage[utf8]{inputenc} 
\usepackage[T1]{fontenc}    
\usepackage{hyperref}       
\usepackage{graphicx}       
\usepackage{url}            
\usepackage{booktabs}       
\usepackage{amsfonts}       
\usepackage{nicefrac}       
\usepackage{microtype}      
\usepackage{xcolor}         
\usepackage{multirow}
\usepackage{tikz}
\usepackage{amsmath}
\usepackage{array}
\usepackage{wrapfig}
\usepackage{subcaption}
\usepackage{xspace}
\usepackage{placeins}
\usepackage{comment}  

\newcommand*\circled[1]{\protect\tikz[baseline=(char.base)]{
            \protect\node[shape=circle,draw,inner sep=1pt] (char) {#1};}}

\newcommand{\algo}{\texttt{M\&C}\xspace}

\newif\ifcomments
\commentsfalse

\ifcomments

\usepackage{soul}

\newcommand{\authorcomment}[2]{\tikz[baseline=(X.base)]\node [draw=#1,fill=#1!40,semithick,rectangle,inner sep=2pt, rounded corners=3pt] (X) {#2};}

\newcommand{\bl}[1]{\authorcomment{blue}{Basile:} \textcolor{blue}{\textit{#1}}}
\newcommand{\lc}[1]{\authorcomment{orange}{Lydia:} \textcolor{orange}{\textit{#1}}}
\newcommand{\rb}[1]{\authorcomment{cyan}{Robert:} \textcolor{violet}{\textit{#1}}}

\else
\newcommand{\authorcomment}[2]{}

\newcommand{\bl}[1]{}
\newcommand{\lc}[1]{}
\newcommand{\rb}[1]{}
\fi

\setcopyright{none}
\acmDOI{}
\copyrightyear{}
\acmYear{}
\acmDOI{}

\acmISBN{}


\title{Match \& Choose: Model Selection Framework for Fine-tuning Text-to-Image Diffusion Models}

\author{Basile Lewandowski}
\authorsaddresses{}
\author{Lydia Y. Chen}
\authorsaddresses{}
\affiliation{%
  \institution{University of Neuchâtel}
  \country{Switzerland}
}
\author{Robert Birke}
\authorsaddresses{}
\affiliation{%
  \institution{University of Turin}
  \country{Italy}
}

\begin{document}

\begin{abstract}
Text-to-image (T2I) models based on diffusion and transformer architectures advance rapidly. They are often pretrained on large corpora, and openly shared on a model platform, such as HuggingFace. Users can then build up AI applications, e.g., generating media contents, by adopting pretrained T2I models and fine-tuning them on the target dataset. 
While public pretrained T2I models facilitate the democratization of the models, users face a new challenge: which model can be best fine-tuned based on the target data domain? Model selection is well addressed in classification tasks, but little is known in (pretrained) T2I models and their performance indication on the target domain. In this paper, we propose the first model selection framework, \algo, which enables users to efficiently choose a pretrained T2I model from a model platform without exhaustively fine-tuning them all on the target dataset. The core of \algo is a matching graph, which consists of: (i) nodes of available models and profiled datasets, and (ii) edges of model-data and data-data pairs capturing the fine-tuning performance and data similarity, respectively. We then build a model that, based on the inputs of model/data feature, and, critically, the graph embedding feature, extracted from the matching graph, predicts the model achieving the best quality after fine-tuning for the target domain. 
We evaluate \algo on choosing across ten T2I models for 32 datasets against three baselines. Our results show that \algo successfully predicts the best model for fine-tuning in 61.3\% of the cases and a closely performing model for the rest.


\end{abstract}

\maketitle

\thispagestyle{plain}
\pagestyle{plain}

\begin{figure}[t]
\includegraphics[width=0.4\textwidth]{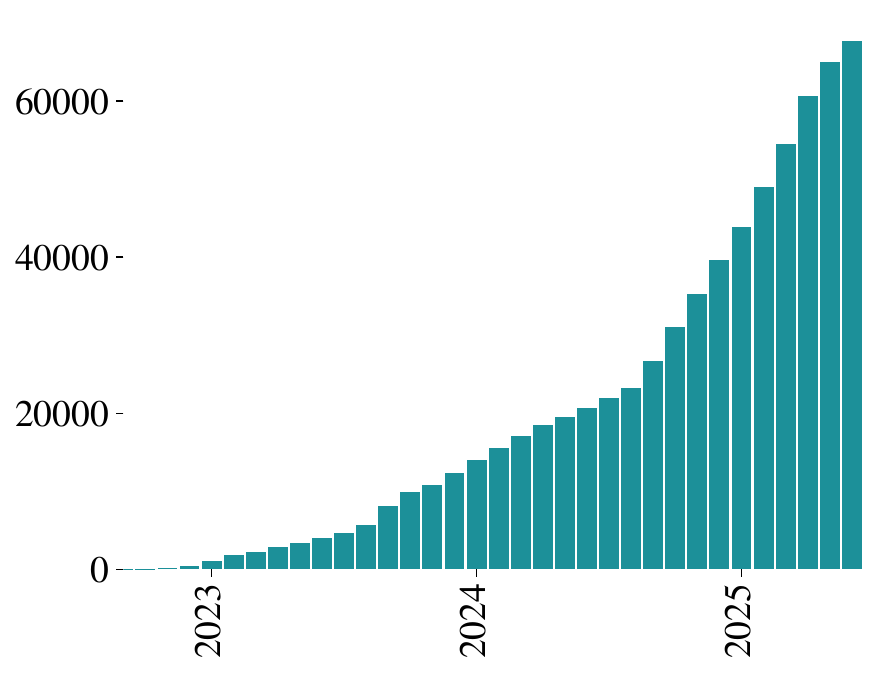}
    \caption{The number of Text-to-Image models publicly available on HuggingFace over time. Source: HuggingFace Hub~\cite{hfapi}.}
    \label{fig:model_nb} 
\end{figure}

\begin{figure*}[htp]
    \centering
 \begin{subfigure}{0.5\textwidth}
 \includegraphics[width=\textwidth]{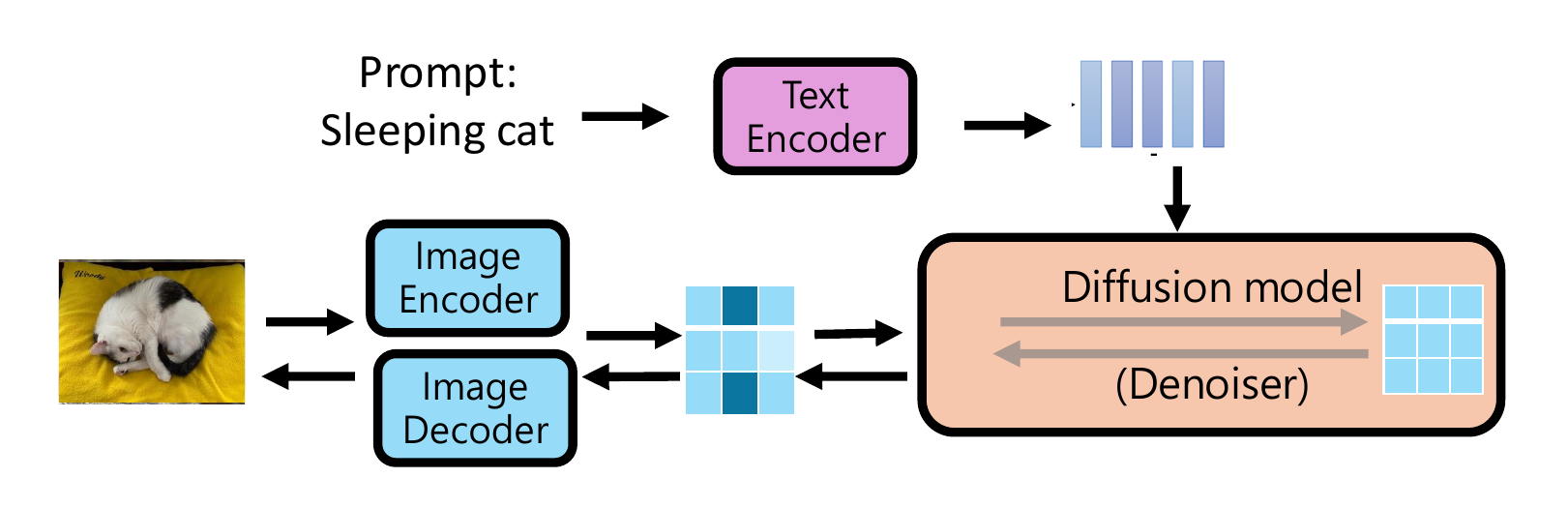}
     \caption{T2I architecture and fine-tuning flow: text decoder, image decoder/encoder, and latent diffusion on the combined image and text latent.  } 
  \label{fig:T2I_FineTune}
  \Description{}
  \end{subfigure}
 \hfill
  \begin{subfigure}{0.45\textwidth}
    \includegraphics[width=\textwidth]{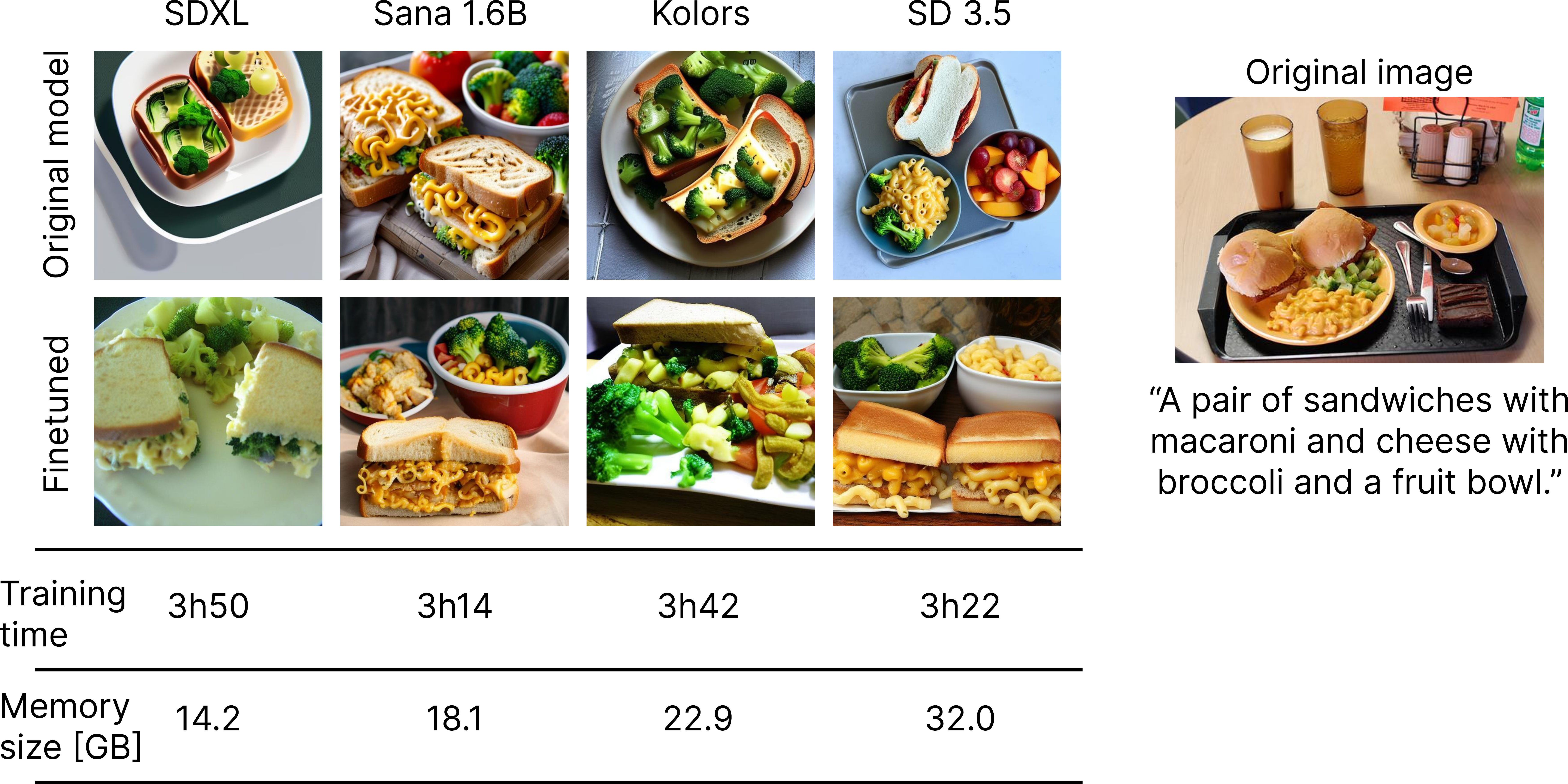}
    \caption{Four T2I fine-tuning overhead (training time) and performance.}
     \label{fig:T2I_Perf}
     \end{subfigure}
  \caption{Text-to-Image fine-tuning: the process, overhead, and the performance.}\label{fig:T2I}
  \Description{}
\end{figure*}

\section{Introduction}
An exceeding number of pre-trained text-to-image foundational models are accessible on open platforms, such as HuggingFace~\cite{hfapi}. Figure~\ref{fig:model_nb} shows the number of such T2I models uploaded to HuggingFace between 2023 and 2025, showing exponential growth.
The mainstream models are latent diffusion models and vision transformers~\cite{10377858, sd15}, which use the embedding of a text prompt to conditionally synthesize images accordingly. They are pre-trained on a large corpus of text-image pairs~\cite{lin2014microsoft}, whose origin is typically unrevealed. 
Users can download such models and adapt to their application scenarios via fine-tuning, i.e., further optimizing the pre-trained model weights using a small amount of target data set. For instance, users can adopt the pre-trained PixArt $\Sigma$~\cite{pixart}, a transformer-based diffusion model, to generate animal images based on prompts and the styles of their target images by further training the downloaded PixArt $\Sigma$ model. 

While model platforms greatly ease the hurdle of accessing pre-trained models, a new challenge arises: which model to choose from the model zoo of T2I for fine tuning given the target images. The naive approach is to download each model and fine-tune all on the target data. This tuning process requires storing the pre-trained models locally and training such models over the target data. The training overhead scales with the number of models considered for any given target dataset. 

Figure~\ref{fig:T2I} illustrates the fine-tuning process, and storage and computation overhead of T2I models, and the performance. Specifically, on Figure~\ref{fig:T2I_Perf}, an example of generating images based on the prompt of ``A pair of sandwiches with macaroni and cheese with broccoli and a fruit bowl.", using four T2I models, namely, Stable Diffusion XL, Santa 1.6B, Kolors, and Stable Diffusion 3.5. We present the images generated from the pre-trained models from the model platform, fine-tuned models using target datasets, the corresponding model sizes, and the fine-tuning time. The evaluation details can be found in Section~\ref{sec:eva}. We observe that the fine-tuned models improve the quality of generated images significantly. At the same time, fine-tuning those models incurs non-negligible overhead, and the model sizes are not a clear indicator for fine-tuning quality. This example highlights the need for a lightweight solution to select a pre-trained T2I model for fine-tuning, but also the complexity of choosing the right model, varying across datasets.

The model selection problem has been addressed for image classification. For a given classification task and target data set, how do we choose a model from the model zoo such that the performance of training such a model on the target data is optimized? Existing solutions select a model either for direct use~\cite{ModelSpiderNeurIPS23, GRAF} or for further fine-tuning. 
When choosing the pretrained classifier for further fine-tuning, the dependency among pretrained classifier, target dataset, and fine-tuned classifier needs to be captured. Bayesian optimization~\cite{arango2024quicktune} and graph modeling~\cite{GRAF} are mainstream approaches to minimize the searching space for the best model. For instance, TransferGraph~\cite{TransferGraph} graphically models the dependency of pretrained models and target data, then trains the predictive model for recommending classifiers through the graph embeddings.

However, the aforementioned classifier selection methods are not applicable to T2I generation because classifiers are evaluated faster than generative models, i.e., a small sample is sufficient for classifier evaluation but not for assessing the quality of synthetic images\footnote{Fr\'{e}chet inception distance (FID) is the de-facto metric to evaluate the quality of synthetic images. A lower FID indicates a higher quality.}.
Overall, the model selection is little addressed for generative models, except~\cite{banditFD} that online selects the best pre-trained model for a given prompt, without exploring the fine-tuning scenarios.

In this paper, we propose the first model selection framework, Match\&Choose (\algo{}), for fine-tuning text-to-image diffusion models. \algo predicts and chooses the best fitting pretrained model to fine-tune on the target dataset by matching to the prior knowledge. The core of \algo is the graphical representation -- matching graph of fine-tuned T2I models and profiled datasets, each of which represents a node in the graph. Two types of links, i.e., model -- data, and data -- data, represent the performance dependency between models and the data, and the data similarity, respectively.  \algo combines the T2I model, data, and more critically graph embedding features to train a predictive ranking model. Upon receiving a target data set, \algo online matches the new data set into the matching graph and predicts the best pre-trained model for further downloading and fine-tuning. We evaluate \algo on choosing ten T2I models for 32 datasets against three baselines selection solutions, i.e., naive classifier, pre-trained selection, and fixed model selection. Our results show that we manage to correctly identify the best model in 61\% of the datasets we consider, with our set of selected models performing better than any given model on average.

Our key technical contributions are summarized as follows:
\begin{itemize}
\item We propose  \algo{}, the first framework for selecting pre-trained T2I models for fine-tuning. It eliminates the staggering search overhead for exhaustive fine-tuning T2I models on target data. 
\item The novel construction of match-graph, which encapsulates the model features, data features, and more critically, the model-data and data-data dependency through the graph embedding features.
\item Based on the tuple of model, data, and graph features, a lightweight tree predictive model to select the best-matched pre-trained model for any target data.
\item Our evaluation results on 10 pre-trained models and 32 datasets show that \algo effectively matches the target data and selects the best model to use in 61.3\% of the cases and close-calls in other cases (outscore to optimal of only 3.9).
\end{itemize}

\section{Background and Related Studies}
In this section, we first explain the process, overhead, and performance of fine-tuning T2I models. Then we summarize the state-of-the-art model selection methods, which center around classification models. 

\subsection{Text-to-Image Generation}

\textbf{Diffusion Models}~\cite{10.5555/3045118.3045358, DDPM}, generate images by approximating the reverse of a stochastic process transforming data to noise, have become the standard approach for text-to-image generation~\cite{dhariwal2021diffusionmodelsbeatgans}. Diffusion models modify data, e.g., images, through two processes, namely the forward and backward processes. The forward process consists of transforming data into pure Gaussian noise by gradually adding Gaussian noise across multiple diffusion steps. At the same time, the backward process aims to gradually restore the noisy data back to a clear image by training a denoiser. Models typically considered for fine-tuning fall into either the U-Net~\cite{unet} or Image Transformer (ViT)~\cite{10377858} categories. The trained denoiser is then used to generate images from random Gaussian noise.

\textbf{Text-to-Image Diffusion} Image diffusion models are then extended to text-to-image models, which first represent the text and images into a latent space and perform forward-backward diffusion to train the denoiser as shown in Figure~\ref{fig:T2I_FineTune}. The text prompt is first mapped into a text embedding through a text encoder that is usually pretrained on a large corpus. The images are also first compressed into compact latent representations. The diffusion process is then applied to the combined latents of images and text prompts. Afterward, the decoder is used to revert the latent back to the image space, which contains the semantic meaning of the text prompts. 

\textbf{Fine-tuning T2I Models} Various T2I models are pretrained on a large corpus of text-image data and are openly shared on the hosting model zoo. Those pretrained models provide reasonable performance on general images. When users have specific requirements for the generated images, e.g., related to style or genre, users can fine-tune one of the pretrained models on the target data. In fine-tuning, one needs to further train the T2I model using the text-image pairs of the target dataset. 
Typical fine-tuning only considers the diffusion model in the image generation pipeline, leaving the other parts (text encoding, image pre-processing) in their initial state. Training is usually focused on the larger layers of the models (e.g. Attention blocks) as they greatly impact the output of the model.
There exist several fine-tuning approaches~\cite{hu2022lora, ruiz2023dreambooth, 52799}, which can accelerate this process. We deem them orthogonal to this study, and they can be combined straightforwardly with the proposed solution. Figure~\ref{fig:T2I_Perf} shows an example of generating the image based on the prompt of ``A pair of sandwiches with macaroni and cheese with broccoli and a fruit bowl.'', using four T2I models, namely Stable Diffusion XL~\cite{sdxl}, Sana 1.6B \cite{ssana}, 
Kolors~\cite{kolors} and Stable Diffusion 3.5~\cite{sd35}. We generate images from the pretrained model (without fine-tuning) and fine-tuned models.
From this small example of four T2I models, fine-tuning visibly improves image quality, but unfortunately, can easily take up to 32GB storage space and multiple hours for each base model considered.

\subsection{Prior Art in Model Selection}

Most previous works on model selection only consider classification tasks. While some can be applied to generative tasks, T2I generation works significantly differently. One of the few works that considers model selection for T2I is~\cite{banditFD} that uses a contextual bandit algorithm to select the best pretrained model without considering fine-tuning.

\textbf{Selection through Knowledge Transfer} Many works have tried to estimate the performance of a model on an untrained task in the context of Transfer Learning~\cite{Tan2021CVPR, TransferTask, 9009545, azizzadenesheli2018regularized, a1140a3d, 10377311}. Some have been successfully applied to model selection. $\mathcal{N}$-LEEP~\cite{NLEEP} substitutes the output layer to perform mutual information estimation. LogMe~\cite{LogME} estimates marginalized likelihood to obtain a proxy for transferability. \citet{NEURIPS2019_e94fe9ac} builds their evaluation on the gradient of few-steps backpropagation. These methods require processing at least a few samples through each considered model, which can become time-consuming and resource-intensive, especially given a large set of models.

\textbf{Selecting Pre-trained Models} In the context of pretrained model selection, ModelSpider~\cite{ModelSpiderNeurIPS23} uses token representations of both the models and tasks to sort the different models, further evaluating the top-ranked ones on a specific task. 
Graph-based approaches have also gained traction: GRAF~\cite{GRAF} formulates Neural Architecture Search using a graph representation to enhance interpretability.

\textbf{Selecting pre-trained Models for Fine-tuning}
Recent efforts in automating model and hyperparameter selection have explored a variety of strategies to address the challenges of dynamic setup and efficient transfer learning. \citet{arango2024quicktune} proposes QuickTune, a framework that jointly performs algorithm selection and hyperparameter optimization using Bayesian optimization, switching towards the best models during the fine-tuning process.
Similarly, \citet{deshpande2021linearizedframeworknewbenchmark} introduces a linearized framework (Label Feature/Gradient Correlation) that leverages prior knowledge to identify suitable models for fine-tuning. 
TransferGraph~\cite{TransferGraph} further advances this direction by applying graph learning techniques to select the most promising models for fine-tuning, demonstrating strong performance across diverse tasks.

These works highlight how the growing variety of models calls for efficient model selection methods. However, they are not applicable to T2I generation because they use a per-sample evaluation metric (classification accuracy or similar) for which there is no straightforward replacement when evaluating the quality of generated images. See Section~\ref{ssec:pb_stat} for further discussion.

\section{Analysis and Problem}

\begin{figure*} [htp]
    \begin{subfigure}{.28\textwidth}
        \includegraphics[width=\textwidth]{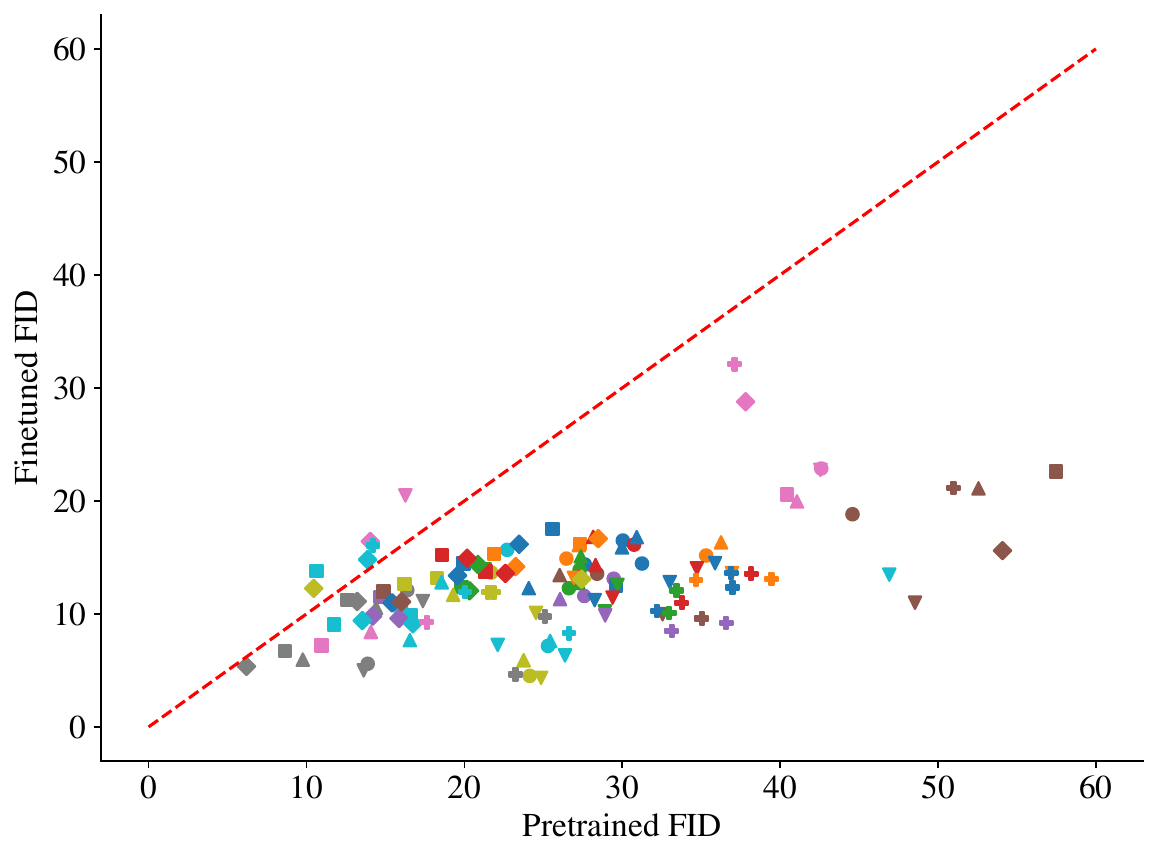}
        \caption{Model Performance (CLIP-FID) before and after fine-tuning (lower is better). Points below the identity line indicate that fine-tuning improves model performance.} 
         \label{fig:finetuning_benefit}
    \end{subfigure}
    \hfill
    \begin{subfigure}{.7\textwidth}
         \includegraphics[width=\textwidth]{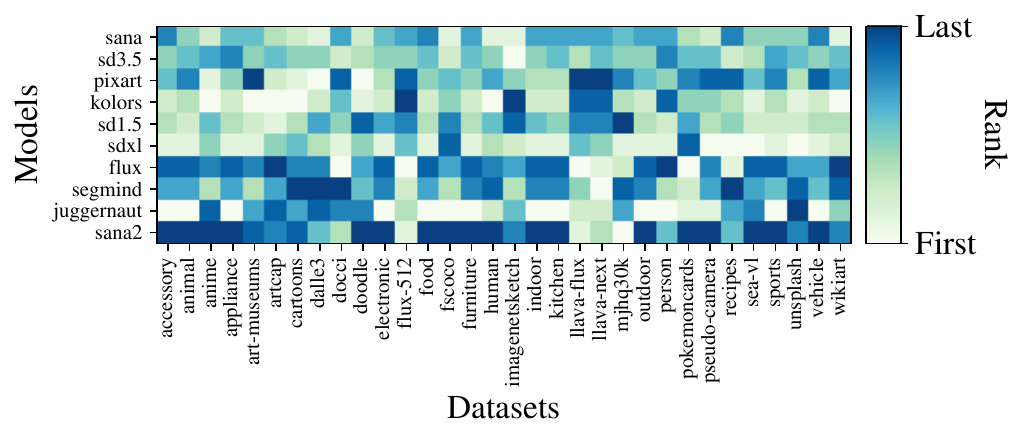}
         \caption{Clip-FID of fine-tuning T2I model on target datasets. Lighter color is better.}
         \label{fig:model_zoo_performance}
     \end{subfigure}
     \caption{The necessity and the performance variation on fine-tuning different models on a given target data set.  }
     \label{fig:analysis}
     \Description{}
\end{figure*}
Before formally defining and addressing the model selection problem, we first investigate its relevance, guided by two key questions: Q1) Can fine-tuning improve quality? and Q2) Can one model outperform all others on all datasets?
The former motivates the benefits and necessity of fine-tuning, the latter excludes a trivial solution to the problem.

\subsection{Benefit of Fine-tuning} \label{ssec:benefit}

The model zoo contains base models trained on typically huge but generic datasets. Fine-tuning of a T2I model is the process that adapts the weights $\theta$ of such a pre-trained base model to the generation task of a specific dataset $\mathcal{D}$.
The dataset $\mathcal{D}$ consists of paired data samples ($x \sim \mathcal{X}, y \sim \mathcal{Y})$, where $x$ represents the textual input (e.g., prompts), and $y$ denotes the corresponding target image. The function $f$ denotes the model's output image generated from text $x$ under parameters $\theta$. Let  $\mathcal{L}$ be a loss function measuring the discrepancy between the generated image and the ground truth image. Then, fine-tuning can be defined as an optimization problem finding the parameters $\theta$ that minimize the expected loss $\mathcal{L}$ over the training data $\mathcal{D}$:

\begin{equation}    
\tau: \theta, \mathcal{D} \mapsto \underset{\theta}{\operatorname{argmin}} \underset{x,y \sim \mathcal{D}}{\mathbb{E}}\left[\mathcal{L}(f_\theta(x),y)\right]
\end{equation}

That is, we adapt the model weights to the target dataset $\mathcal{D}$.  
However, the loss and quality metric functions differ in all practical cases, with no simple relationship between them that we can leverage to predict the model quality performance after fine-tuning.
Hence, to answer R1, Figure~\ref{fig:finetuning_benefit} plots the generated image quality (based on CLIP-FID cf. Section~\ref{ssec:graph-construction}) before versus after fine-tuning for different datasets (represented by different colors) and different models (represented by different point shapes). The lower the CLIP-FID score, the higher the quality of generated images. As expected, most points lie below the diagonal line, indicating that fine-tuning improved the model's performance. The figure also underlines that different model-dataset combinations achieve different performance gains, and performance even decreases for a tiny portion of models. 
However, these are never the best model for the given dataset, and overall, we can answer Q1 that fine-tuning indeed improves quality.

\subsection{No Trivial Solution} \label{ssec:one_fits_all}

Having assessed that fine-tuning benefits generation quality to different degrees, one could wonder if there is no trivial solution to the model selection problem. If there exists a model that achieves the best performance across all or most datasets, that model would be a trivial solution to our problem. Figure~\ref{fig:model_zoo_performance} shows that this is not the case. The heatmap plots the generation quality after fine-tuning for all combinations of datasets (on the x-axis) and models (on the y-axis). The lighter the color, the higher the model's performance. While some models, e.g., Kolors or SDXL, obtain on average better quality across the datasets, none are exempt from problematic darker spots. Moreover, the darker spots do not always correspond to the same datasets.
For example, Kolors and SDXL have almost opposite results for datasets like Anime, Cartoons, FS-COCO, Imagesketch, and Pokemoncards.
This answers Q2, underlining that a trivial one-model-fits-all solution does not exist and motivates the importance of model selection for achieving the best results.

\subsection{Problem Statement} \label{ssec:pb_stat}

Model selection is explored mainly in the context of classification tasks. Critically, such tasks can be evaluated on a few model outputs using universally accepted objective metrics, e.g., classification accuracy. 
Generative T2I models lack such measures. This renders previous methods inapplicable.
Evaluating image generation is a harder and more computationally intensive problem.
The standard approach to obtain a reliable quality estimation is to evaluate the discrepancy between a fine-tuned model's output and real data.
One such measure is the Fr\'{e}chet Inception Distance (FID)~\citep{FID}, a metric that computes the distance between generated and real images by embedding both into a lower-dimensional feature space using a pretrained model (typically the Inception network~\citep{Szegedy_2016_CVPR}, hence the name). The score is then computed as the Fr\'{e}chet distance~\citep{frechet} between the embeddings. In our case, we will use the CLIP~\cite{CLIP} model for image embeddings by default, as suggested by \citet{Kynkaanniemi2022}.
This complex evaluation requires significantly higher compute resources and makes it difficult to predict, as there are no obvious correlations between the model structure/parameters and the outcome of fine-tuning or the commonly used quality scores, e.g., FID.
Here, we aim to build a model selection framework that can overcome both problems. This framework leverages known model performances and dataset similarities to predict the goodness of fine-tuning a model on a new dataset without actually doing it.

\section{Match and Choose \algo } \label{sec:algo}

The difficulty in model selection for text-to-image generation stems from the large sets of samples required by the quality evaluation metrics to assess a model’s performance. Given our goal of building a lightweight selection framework, we propose leveraging prior knowledge of each model's capability to avoid costly evaluations. We will hereafter introduce our prediction framework, \algo, which first builds a \textit{matching graph}. Then we describe how we represent prior knowledge with it and use it to train a ranking model. Finally, we explain how to use it to predict the performance of a new T2I model without explicitly fine-tuning it first.

\subsection{Overview}

\begin{figure*}[t]
\includegraphics[width=\linewidth]{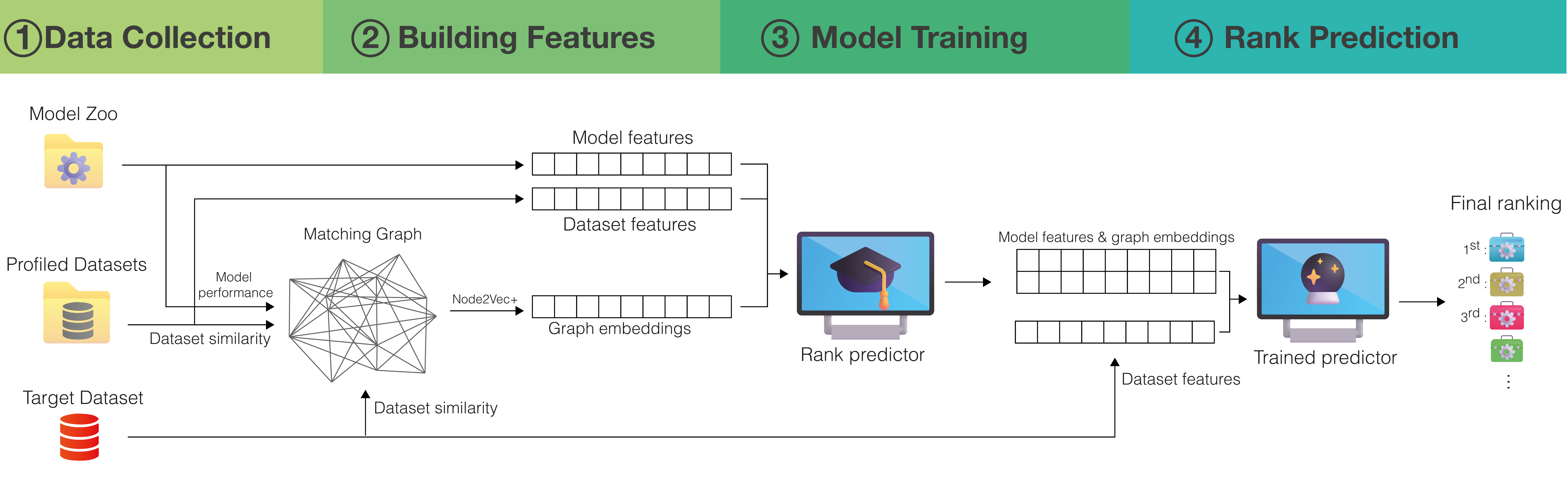}
\caption{Overview of \algo. Steps \circled{1}, \circled{2} and \circled{3} train the ranking model offline. Step \circled{4} predicts the best model for a new dataset.}
\Description{}
\label{fig:overview}
\end{figure*}

We propose \algo, shown in Figure~\ref{fig:overview}, to enable reliable model selection and performance prediction across datasets.
\algo encompasses four main steps: three offline and one online.
The key to \algo is a \textit{matching graph} built in Step \circled{1}. This graph encapsulates model performance and dataset similarity into a single data structure. 
It consists of two types of vertices —datasets and models— and two types of edges: one representing how well a model performs on a dataset, and the other reflecting the similarity between datasets. Both edge types are weighted using the Fr\'{e}chet Inception Distance (cf. \autoref{ssec:pb_stat}), which allows
to represent 
model performance and similarity relations 
via the same metric.

Depending on its type, each graph vertex incorporates feature representations for a dataset or model extracted in Step \circled{2}.
Model features are derived from their configuration metadata, and dataset features by averaging embeddings obtained via a probe model over the images in each dataset, capturing their different characteristics. 

With this enriched representation, we formulate a prediction task: rank models based on their expected performance on a target dataset. Using the matching graph,
Step \circled{3} trains a model to rank the different T2I models of our zoo. This model is trained offline on known performance rankings. Finally, Step \circled{4} predicts the model rankings using the trained model's ability to generalize to new datasets by estimating the relative effectiveness of each candidate model without requiring prior fine-tuning results. We detail each step in the next four subsections.


\subsection{Model and Dataset Features}
We extract features describing the different models and datasets as input for our prediction algorithm. These features are conceptually associated with the vertices of the matching graph. 
The purpose of these features is to base our prediction not only on the relations between models and datasets, but also the particular characteristics of the latter.

\textbf{Model Features} We extract model features using the metadata from the multiple T2I pipelines that we consider. We parse the models' configuration files and retain the 
hyperparameters that differ globally across the whole zoo. 
We also add two values not present in the configuration files: the number of parameters and the throughput (in FLOPS) of the pipeline on a test workbench. Finally, the result is processed through normalization and one-hot encoding to obtain our model features.  In the end, we consider 78 different pipeline hyperparameters (see Table~\ref{tab:features} for examples) and map each model to a 165-dimensional embedding space. 

We report in \autoref{tab:features} the features impacting the model output the most. The features related to transformer concern the models that use a ViT architecture, whereas the U-Net ones are related to the U-Net models. Throughput and number of parameters are among the most important features, suggesting that they are indeed interesting indicators of a model's performance. However, they are not significant enough to generate a prediction on their own, as they remain in small proportion to the majority.


\begin{table}[t]
    \centering
    \caption{Relative importance of the Different Model Features. We report the random forest's change in prediction value for the different model features we consider and report the top 10 ones.}
    \begin{tabular}{lc}
    \toprule
    Feature & Importance \\
    \midrule
    \verb|throughput (flops)| & 14.5 \% \\
    \verb|transformer_num_layers| & 4.4 \% \\
    \verb|unet_num_params| & 4.4 \% \\
    \verb|transformer_attention_head_dim| & 3.6 \% \\
    \verb|transformer_sample_size| & 2.8 \% \\
    \verb|vae_sample_size| & 2.6 \% \\
    \verb|unet_attention_head_dim_1| & 2.5 \% \\
    \verb|unet_block_out_channels_2| & 2.3 \% \\
    \verb|unet_transformer_layers_per_block_2| & 2.2 \% \\
    \verb|unet_mid_block_type_UNetMidBlock2DCrossAttn| & 2.1 \% \\
    \verb|Remaining| & 58.5 \% \\
    \bottomrule
    \end{tabular}
    \label{tab:features}
\end{table}

\textbf{Dataset Features} We extract dataset features via a process similar to previous works~\citep{LogME,ModelSpiderNeurIPS23,TransferGraph,Cui2018iNatTransfer}. The idea is to process all the images through the same probe model for each dataset and then average the embeddings produced. In classification model selection settings, the probe model is often ResNet or unsupervised classification, sometimes with weighted averaging by class. In our setup, we considered different models for this task: Sentence-T5~\citep{ni-etal-2022-sentence} for caption parsing, ImageBind~\citep{girdhar2023imagebind} for captions and images, and CLIP~\citep{CLIP} for images only. We obtain the best result using CLIP cf. \autoref{sec:alt_feats}.

\subsection{Matching Graph Construction} \label{ssec:graph-construction}

To make reliable predictions, we need a way to extract the knowledge from the data we already have for both datasets and models.
Therefore, we design a \textit{matching graph} to represent two relations: i) how a model performs on a dataset, and ii) how similar two datasets are. The resulting graph 
contains two types of vertices: one for datasets and one for models. 
The graph also contains two types of edges. 
Edges between models and datasets represent model performance. Edges between datasets represent dataset similarity.
Edges between models are discarded as they do not provide valuable information.

As edge weight, we use the Fr\'{e}chet distance \(d_{F}\)~\citep{frechet} between embeddings. This approach's novelty is the proposal to leverage \(d_{F}\) to measure similarity between datasets and model goodness. Given two image embedding distributions $(\mathcal N(\mu, \Sigma)$ and $\mathcal N(\mu', \Sigma')$, \({d_{F}^2}\) is defined as:

\begin{equation}
    d_{F}^2 = \lVert \mu - \mu' \rVert^2_2 + \operatorname{tr}\left(\Sigma + \Sigma' -2\left(\Sigma \Sigma'  \right)^\frac{1}{2} \right)
\end{equation}

Let $\psi: \mathcal Y \rightarrow \mathbb R^d$ denote an embedding model (e.g., Inception), which maps an image into a $d$-dimensional feature space, we define our evaluation function as

\begin{equation} \label{eq:eval}
    \phi: f, \mathcal{D} \mapsto d_{F}\left( \psi\left(f(\mathcal D)\right), \psi\left(\mathcal D\right) \right)
\end{equation}
We consider FID as our default evaluation function $\phi$ for the rest of this article, using CLIP~\citep{CLIP} as the embedding model.





We denote \( G = (V, E, \phi) \) as our matching graph, where \( V \) is a set of vertices, \( E \subseteq V \times V \) is a set of edges, and \( \phi \) is the weight function that assigns a real-valued weight to each edge, independent of its type. In our case, $\phi$ can evaluate a model's performance as defined in \autoref{eq:eval} or measure similarity between datasets FID, depending on the type of the edge.

The matching graph's main purpose is to store model performance and similarity values. 
We define the edge weights as follows.\\
\textbf{Model – Dataset Edge} We compute the FID score of the model that has been fine-tuned on the linked dataset and assign this value to the edge. By doing so, the edge effectively reflects how well the model was trained on this particular dataset. \\
\textbf{Dataset – Dataset Edge} The edge weight is the FID between the two datasets. As the FID measures the distance between two distributions, it can be used to calculate the discrepancy between unrelated datasets.

Consequently, even if the matching graph itself is heterogeneous with two types of vertices, the edge weights are computed using the same metric. This enables us to streamline the representation process in subsequent steps, as it can be interpreted as a homogeneous undirected graph when only the edges are considered.

\subsection{Training Ranking Model}
\begin{figure}
    \centering
    \includegraphics[width=\columnwidth]{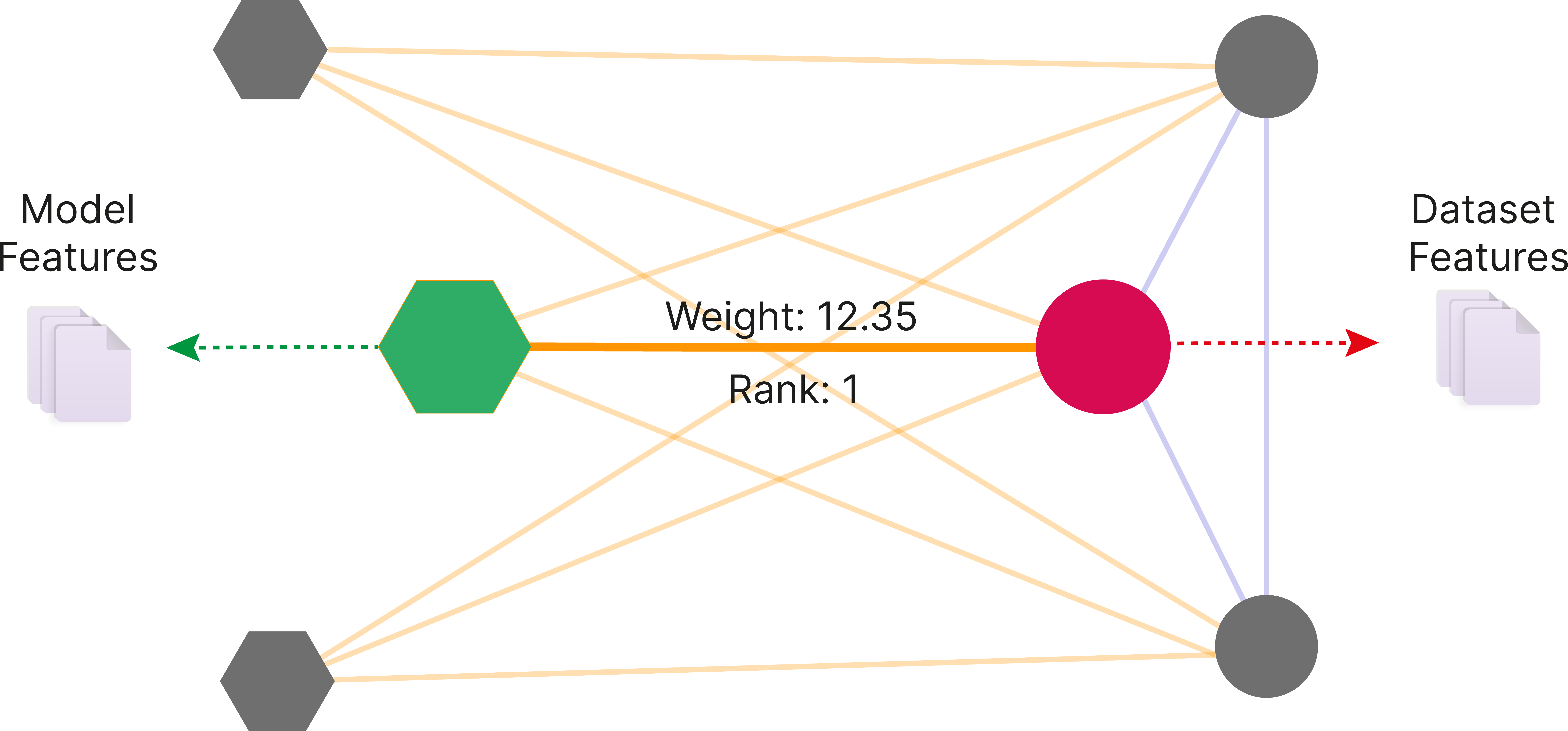}
    \caption{Example of Edge Value. This model -- dataset edge value is given by the FID score of this model fine-tuned on this dataset. Once the graph is constructed, we also consider its rank, i.e., how this score compares to the scores of the other models connected to this dataset. The training objective of our estimator is to accurately predict this rank.}
    \label{fig:edge}
    \Description{}
\end{figure}



Steps \circled{1} and \circled{2} set up the matching graph. Step \circled{3} uses it as input to train a model to predict the ranking of the different models on a target dataset. This rank reflects each model's performance after being fine-tuned on the target dataset.

First, Step \circled{3} processes the matching graph such that we can learn from it. 
It starts by computing an embedding for each edge using Node2Vec+~\citep{10.1093/bioinformatics/btad047}. Node2Vec is a graph embedding algorithm that performs random walks through the graph, guided by the weight of the edges. The embeddings it creates provide a representation of the proximity of one node to its neighbors and an indication of the similarity between regions of the graph.

Our goal is to predict the value of the edges between models and datasets, since they represent the models' performances. Given the edge embedding extracted from the graph via Node2Vec+, we concatenate it with the features of the two connected nodes and use it as input for the ranking model.
Therefore, one input of our predictor is the concatenation of three vectors: the model features, the dataset features, and the edge embedding.
Further, we rank models based on their performance for each dataset. We then assign each model's rank to the corresponding model-dataset edge. This rank is used as the target value for our predictor. 


These input-rank pairs serve as the training set for a classifier model whose task is to predict each model's rank.
Each rank is considered a separate class. Based on preliminary experiments, we use CatBoost~\citep{catboost} to learn the predictions for the best accuracy. We also report the results of using other models as predictors in \autoref{sec:alt_pred}.


\subsection{Predicting T2I Model Rank}


Once we have a trained classifier, Step \circled{4} forecasts how the models will perform on a new dataset, starting by assembling the concatenated input embedding. 
As for the training process, this input consists of the features from the models and the target dataset, as well as a representation of the edge in the graph. Here, we must compute the dataset features and similarity edges with other datasets. 
Instead, the model features are the same as during training. Then, we compute the edge embedding via Node2Vec+ without considering the edges from the dataset to the models, as we do not know their weight beforehand.
Finally, we concatenate the three vectors for each model in the zoo and feed the result into the classifier to estimate the rank. This rank represents the relative performance of the model after being fine-tuned on the target dataset. By picking the top-ranking model, we can select the T2I model to fine-tune to achieve the highest generation quality.


\section{Experimental results }\label{sec:eva}
In this section, we aim to evaluate the performance of \algo{} in selecting the best-fitted pre-trained model for further fine-tuning. 
We consider metrics related to image quality and text-image correspondence. We show the effectiveness of \algo{} by the relative difference of those metrics against baselines.
Prior to showing the results, we first detail the experimental setup, including the models, target dataset, the fine-tuning process and the key parameters. 

\subsection{Evaluation Setup} \label{ssec:setup}

\textbf{Datasets and Models}
We consider in total 32 datasets and 10 pretrained text-to-image models (cf. \autoref{fig:model_zoo_performance})\footnote{The pretrained models and datasets are publicly available on \url{https://huggingface.co/}.}. The datasets consist of captioned image sets of various kinds, including but not limited to the 12 super-classes of the MS-COCO dataset~\cite{lin2014microsoft}. For the datasets where it is required, we crop the images to a $512\times512$ pixels resolution.

In our this setup, the ranking of the models on their average performance is (best to worst): SDXL, SD 3.5, Juggernaut, Sana 1.6B, Kolors, SD 1.5, Pixart, Segmind Vega, Flux dev, Sana 4.8B

\textbf{Setup} 
We fine-tune each model for 10~000 steps with a batch size of 8 and a learning rate of $10^{-5}$ on AdamW optimizer. As is commonly the case with fine-tuning, we use Low Rank Adapters (LoRA)~\citep{hu2022lora} to streamline the training process. This approach involves training a low-dimensional replica of the layer rather than the entire network, thereby reducing the required computing resources. We keep the text encoder and feature extractor models as provided and use their original parameters throughout the training and evaluation.

We run the fine-tuning on two Nvidia H100 GPUs and an Intel Xeon Platinum 8562Y+ CPU, using PyTorch 2.6 and CUDA 12.8 on Ubuntu 22.04.5. Fine-tuning takes 2 to 8 GPU hours depending on the model and target dataset, which is in any case orders of magnitude smaller than the initial training cost of the models we consider.

We evaluate each model on a validation set of 10~000 samples and compute the FID as described in \autoref{ssec:pb_stat} using the cleafid library~\cite{parmar2022aliased}. We therefore compute the ground-truth value for all the model -- dataset pairs considered in our evaluation.

\autoref{tab:models} lists each of our model's main features. These models reflects a large variety in architectures and original training datasets. We used bf16 floating point arithmetic for Kolors and Sana v1.5 and fp32 for the other models. Throughput and sampling speed were measured on the same platform we used for training.



\begin{table}[]
    \centering
    \caption{Pretrained base models. Sampling time is to generate 1 image.}
        \begin{tabular}{lrrrr}
            \toprule
             Model & Year & TFLOP & \parbox[c]{1cm}{Param. \\ $[\text{G}\#]$} & \parbox[c]{1cm}{Sampling\\time [s]}\\
             \midrule
            Flux.1 dev~\cite{flux} & 2024 & 563.25 & 11.90 & 19.90\\
            Juggernaut X~\cite{jugernaut} & 2024 & 155.98 & 2.57 & 5.25 \\
            Kolors~\cite{kolors} & 2024 & 174.35 & 2.58 & 5.81 \\
            Pixart $\Sigma$~\cite{pixart} & 2024 & 52.67  & 0.61 & 2.17 \\
            Sana 1.6B~\cite{ssana} & 2024 & 38.28  & 1.60 & 1.00\\
            SD 1.5~\cite{sd15} & 2022 & 71.59  & 0.86 & 2.42 \\
            SD 3.5~\cite{sd35} & 2024 & 124.42 & 2.47 & 5.30 \\
            SDXL~\cite{sdxl} & 2023 & 155.98 & 2.57 & 5.25\\
            Segmind Vega~\cite{segmind} & 2024 & 67.01  & 0.75 & 1.60 \\
            Sana 4.8B~\cite{ssana2} & 2025 & 360.60 & 4.72 & 8.06 \\ 
            \bottomrule
        \end{tabular}
    \label{tab:models}
\end{table}

\textbf{Metrics} We introduce several metrics to compare the performance of the different model selection frameworks.
\begin{itemize}
    \item \textbf{Optimal Selection Score} (OSR): Proportion of the datasets where the algorithm successfully predicts the best model. In our case it is the proportion over the 32 datasets.
    \item \textbf{Kendall weighted} $\tau_w$: Correlation coefficient defined by \citet{kendalltau} that measures the ordinal association between two ranked variables, giving more weight to disagreements at higher ranks.
    \item \textbf{Outscore to Best} (O2B): difference of the predicted best model's score with best model over all datasets on average. Negative value means the selected models actually perform better than it on average.
    \item \textbf{Outscore to Optimal} (O2O): difference of the predicted best model's score with the actual best model for each dataset. Theoretical best value is 0.
\end{itemize}

\textbf{Baselines} Since, to the best of our knowledge, there is no available method for T2I model selection, we construct three different baselines for comparison with the proposed method. 
\begin{itemize} 
    \item \textbf{Initial}: we report the ranking of the models before they undergo fine-tuning. 
    \item \textbf{Overall}: we always report the ranking over every datasets on average. The ranking is thus the same for every dataset.
    \item \textbf{Direct Classifier}: we train a multi-output random forest classifier using only the dataset features.
\end{itemize}

\subsection{Full Scale Results} 

\begin{table}[]
    \centering
    \caption{Evaluation results on Model Selection.}
    \begin{tabular}{lcccc}
        \toprule
         Method & OSR $\uparrow$  & Kendall $\tau_w$ $\uparrow$ & O2B $\downarrow$ & O2O $\downarrow$  \\
         \midrule
         \algo  & 61.3 \% & 0.51 & -1.62 & 3.91\\
         Direct Classifier  & 3.2 \% & -0.12 & 1.68 & 5.68 \\
         Initial  & 3.5 \% & -0.30 & 9.60 & 24.08 \\
         Overall & 12.9 \% & 0.33 & 0.0 & 3.78 \\
         \bottomrule
    \end{tabular}
    \label{tab:results}
\end{table}

We perform model selection on all of our datasets using a leave-one-out pattern. For each dataset, we run the prediction using the data of all other datasets. We construct a separate graph for each dataset by removing the model edges connected to the dataset being evaluated. The result are averaged over all the datasets we consider. The whole prediction process remains very fast, taking no longer than one minute on our platform in any case. 

The results displayed in \autoref{tab:results} show the proposed method outperforms all baselines. \algo manages to accurately predict the best model in the majority of cases, with our selection obtaining an average score better than using best performing model on average (in this case Stable Diffusion XL). In particular, our approach achieves the best ranking correlation (Kendall $\tau_w$), indicating strong consistency with the ground truth model rankings.

We can identify several factors that lead to an incorrect model rank prediction. First, in datasets where all the results are in a close range, the ranking is heavily affected by variance in the evaluation process, making successful rank predictions difficult. Second, in datasets where generation is strongly impacted by model limitations, such as the maximum number of tokens per prompt for image generation including writings in the image. Third, datasets with synthetic images produce unexpected results, as training on synthetic data can be difficult~\cite{nips/GengHRWLKK24}.

The overall baseline, selecting the average best model, is the strongest among our baseline methods, suggesting that a selection based on prior statistical results is stronger than a naive predictor. As outlined by the results from \autoref{ssec:one_fits_all}, the ranking of models before fine-tuning is a poor indicator of their performance after training. This illustrates that larger models perform well overall but are more difficult to adapt to specific data.

\subsection{Results with Partial Knowledge}

\begin{figure}[t]
    \centering
    \includegraphics[width=\columnwidth]{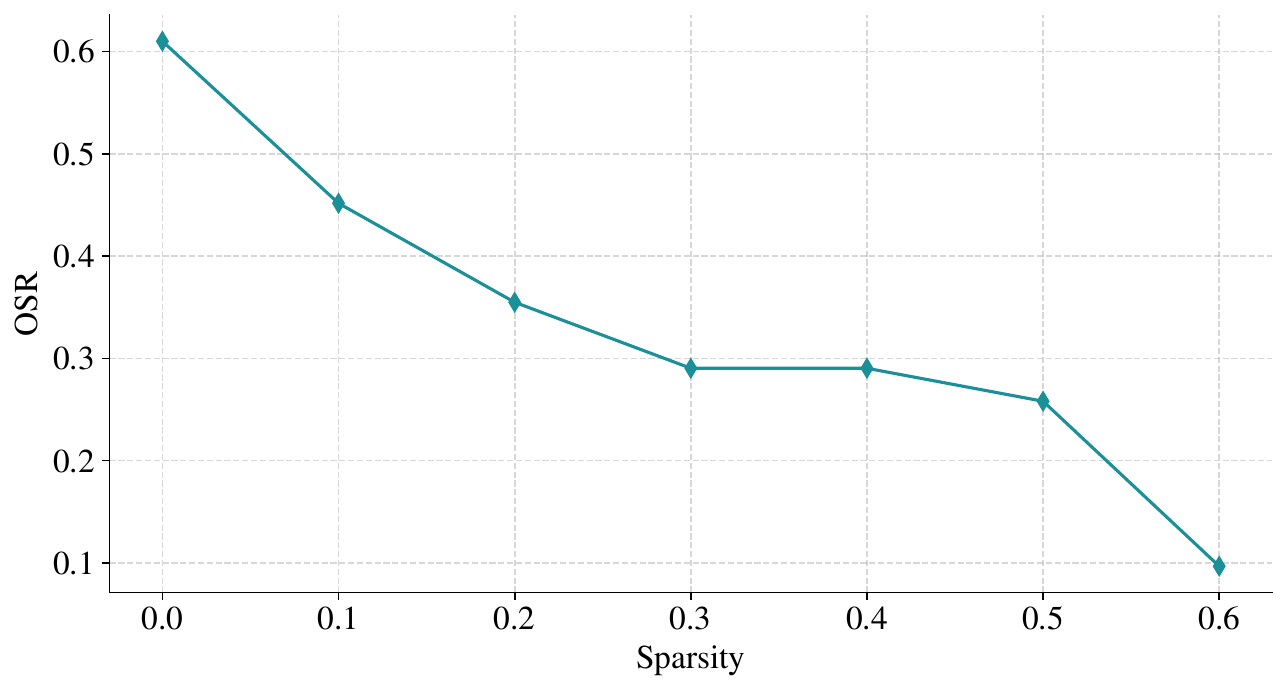}
    \caption{Impact of the knowledge deficit on Optimal Selection Ratio (OSR). The ability to identify the best model decreases as we remove model performance information from the matching graph. With data deleted from 60\% of the model -- dataset pairs, our prediction fails to outperform a random guess.} 
    \label{fig:partial}
    \Description{}
\end{figure}

The previous section uses the fine-tuning results of all models on all datasets, which is supposedly a very densely connected graph. However, this information is only available after fine-tuning all models on the target datasets.
This defeats the purpose of predicting the rank.
Therefore, we study how the accuracy of the prediction is impacted when we assume we only have partial information on the models' performance. We display in \autoref{fig:partial} the evolution of the Optimal Selection Ratio as the amount of information on the models performance decreases. 

A 10\% sparsity means we remove randomly 10\% of the model--dataset edges which represent the model performance information, therefore decreasing the size of the training set 
in the same proportion. (0\% represents the case we considered in the previous subsection.) We observe that removing information degrades the accuracy of the prediction, but \algo maintains some prediction capability until a large part of the information is lost. 
This suggests we need a minimal set of models for which we know the performance on most-to-all datasets. The more information at hand, the better the predictions. If we want to increase the prediction ability, there is also the possibility to enrich the matching graph after we finish fine-tuning the model predicted by \algo on a new target dataset. The drawback is that, if we trust \algo, we only add one model-dataset edge each time.


\subsection{Ablation Study}

\begin{table}[t]
    \centering
    \caption{Evaluation Results with partial input data when forgoing either node or connectivity features.}
    \begin{tabular}{lcccc}
        \toprule
         Input & OSR $\uparrow$ & Kendall $\tau_w$ $\uparrow$\\
         \midrule
         Model \& Dataset features &  48.4 \% & 0.46 \\
         Graph embeddings & 29.0 \% & 0.38  \\
         \midrule
         \algo (both) & 61.3 \% & 0.51  \\
         \bottomrule
    \end{tabular}
    \label{tab:ablation}
\end{table}

The input of our algorithm is of two kinds, as presented in \autoref{sec:algo}. First, the features extracted from the models and datasets (all of which are independent of one another); and, second, the embeddings extracted from the matching graph. Although some information is redundant between these two types of input, we argue that both are necessary because they provide information about each object on the one hand and about the objects' relationships on the other hand.

To support this claim, we run a reduced version of our algorithm with only part of the input data. As shown in \autoref{tab:ablation}, using only the model/data features or only the graph embeddings produces
worse results. 
In particular, using only graph embeddings leads to less than half the OSR score. The Kendall $\tau_w$ follows a similar trend. This indicates that both contribute, but especially model and dataset features are an important source of information.

\subsection{Alternate Quality Metrics}

\begin{table}[t]
    \centering
    \caption{Evaluation results with various metrics in \algo. To enable valid comparisons between methods, we report O2B and O2O as a proportion of their respective value ranges. As such, the evaluation based on ImageBind shows very small values because the computed distances are of high magnitude.}
    \begin{tabular}{lcccc}
        \toprule
         Metric & OSR $\uparrow$ & Kendall $\tau_w$ $\uparrow$ & O2B $\downarrow$ & O2O $\downarrow$ \\
         \midrule
         CLIP-FID  & 61.3 \% & 0.51 & -1.1 \% & 6.8 \% \\
         ImageBind-FID & 56.6 \% & 0.36 & -0.1 \% & 0.1 \% \\
         FID  & 37.9 \% & 0.40 & 1.7 \% & 6.6 \% \\
         KID  & 30.7 \% & 0.30 & 2.4 \% & 4.3 \% \\
         CMMD & 48.3 \% & 0.41 & 0.9 \% & 7.0 \% \\
         \bottomrule
    \end{tabular}
    \label{tab:alternates}
\end{table}

Despite remaining the gold standard for T2I evaluation, FID has two major drawbacks. First, it relies on the assumption that the image embeddings follow a multivariate Gaussian distribution. Second, it requires many samples to provide a reliable estimate. Several studies have attempted to address these issues (such as KID~\citep{BinkowskiSAG18}, $\text{FID}_\infty$~\citep{fidinf}, and CMMD~\citep{cmmd}), but they remain less widely used.

We therefore explore the performance of \algo when using other synthetic image quality measures. In addition to FID, we present results using KID and CMMD: these methods measure the similarity between real and generated images using Maximum Mean Discrepancy (MMD), offering an alternative to Fréchet distance by comparing feature distributions using a polynomial kernel or conditionally, respectively. Moreover, for FID, by default in the rest of the paper, we use the CLIP model for feature extraction, while here we also report the results when using ImageBind and InceptionV3 (as in the original definition) as embedding models.

The results reported in \autoref{tab:alternates} show that we maintain consistent performance on the other distance-based metrics. We can surmise that CLIP embeddings are superior in our case since they are specifically designed for T2I generation with a model similar to those often used in the text encoders of our pipelines.

\subsection{Alternate Features Extractors} \label{sec:alt_feats}

\begin{table}[t]
    \centering
    \caption{Evaluation results when considering different modalities for dataset feature extraction.}
    \begin{tabular}{lcccc}
        \toprule
         Input & OSR $\uparrow$ & Kendall $\tau_w$ $\uparrow$\\
         \midrule
         ImageBind &  48.1 \% & 0.50 \\
         Sentence-T5 & 44.4 \% & 0.48  \\
         CLIP & 61.3 \% & 0.51  \\
         \bottomrule
    \end{tabular}
    \label{tab:other_feats}
\end{table}

We also consider different probe models we can use to construct our datasets features to try to capture properties of different data modalities. 
We used CLIP as a baseline since it was trained on a large number of image-model pairs and has been commonly used as an image feature extractor~\citep{iclr/ShenLTBRCYK22}. We now also consider two other models to account not only for the images of the datasets. Sentence-T5~\cite{ni-etal-2022-sentence} is an adapted Text-to-Text transfer transformer~(T5) model designed for sentence embeddings. We use it to parse the captions of the different datasets, enabling us to define a dataset feature. ImageBind~\cite{girdhar2023imagebind} is a multi-modality model that maps both the text and the images to the same embedding space. 

The results reported in \autoref{tab:other_feats} indicate that the models based on modalities other than images fail to perform as well as CLIP. It can be argued that using an evaluation framework only based on image quality favors an image-only embedding model. We can also advance that, different from CLIP, these models are not designed specifically for the T2I context, and the specificity of image generation prompts may hinder their abilities.

\subsection{Alternate Ranking Predictors} \label{sec:alt_pred}

\begin{table}[t]
    \centering
    \caption{Prediction Result of different estimators. Notable hyper-parameters are: we used a Radial Basis Function kernel for SVC, a hidden dimension of 256 for the MLP and 300 estimators for CatBoost.}
    \begin{tabular}{lcccc}
        \toprule
         Input & OSR $\uparrow$ & Kendall $\tau_w$ $\uparrow$ \\
         \midrule
         SVC &  16.1 \% & 0.35 \\         
         MLP &  45.1 \% & 0.38 \\
         RF & 51.6 \% & 0.44  \\
         XGBoost & 51.6 \% & 0.42  \\
         CatBoost & 61.3 \% & 0.51  \\
         \bottomrule
    \end{tabular}
    \label{tab:predictors}
\end{table}

Predicting the rank of a model in \algo is represented as a classification problem. Therefore, we compare the performance of several common classification algorithms to explore their impact on the prediction accuracy. Support Vector Classifier~(SVC) and Random Forest~(RF) are classic machine learning models that are widely used in various classification context.

Multi-Layer Perceptron~(MLP) is the most basic example of a feedforward neural network, composed of a simple sequence of linear layers. Extreme Gradient Boosting~(XGBoost)~\citep{10.1145/2939672.2939785} and CatBoost~\cite{catboost} are two more recent tree-based ensemble models that implement gradient boosting, where CatBoost natively handles categorical features.

We report in \autoref{tab:predictors} the different results of these models, observing that the three best predictors are all tree-based. RF, XGBoost, and CatBoost all achieve decent results, as is commonly the case in such a classification setting, though CatBoost fares 9.7 points better. Hence, we make the choice to use CatBoost.

\section{Conclusion}

The explosive number of pretrained T2I models on model-sharing platforms poses the new challenge for users on choosing the best one for fine-tuning. Exhaustive trial and error search on the fine-tuning space is expensive in computational and storage resources. In this paper, we propose the first model selection framework, \algo, which maps the target data set into our novel \emph{match-graph} and chooses the best pre-trained model using the predictive tree model. The unique feature of \algo is to model the model-data dependency using the graph embedding of the matching graph and static features of models and datasets. Such a representation well captures the complex dependency among pre-trained models and target data, enabling the simple tree predictive model to achieve optimal model selection. Our evaluation results against the baseline solutions show \algo effectively matches the target data with prior knowledge and chooses the best pre-trained model in the majority of cases, i.e., reaching 61\% OSR. Moreover, we maintain a negative O2B score, indicating that no single base model in our zoo could provide as strong results.

\section{GenAI Usage Disclosure}

Herein, we confirm that there is no conscious use of GenAI for editing this paper. We do use GenAI to generate Figure~\ref{fig:T2I}, as the objective of this paper is to select and fine-tune the text-to-images generative models. It's necessary to evaluate those GenAI models on our proposed framework and present the results.

{
\small
\bibliographystyle{ACM-Reference-Format}
\bibliography{refs.bib}
}

\end{document}